\documentclass{article}

\usepackage{arxiv}

\usepackage[utf8]{inputenc} % allow utf-8 input
\usepackage[T1]{fontenc}    % use 8-bit T1 fonts
\usepackage{hyperref}       % hyperlinks
\usepackage{url}            % simple URL typesetting
\usepackage{booktabs}       % professional-quality tables
\usepackage{amsfonts}       % blackboard math symbols
\usepackage{nicefrac}       % compact symbols for 1/2, etc.
\usepackage{microtype}      % microtypography
\usepackage{lipsum}		% Can be removed after putting your text content
\usepackage{graphicx}
\usepackage[numbers]{natbib}
\usepackage{doi}

\usepackage{booktabs}
\usepackage{multirow}
\usepackage{caption}
\usepackage{colortbl}
\usepackage{amsfonts}
\usepackage{amsmath}
\usepackage{xcolor}
\usepackage{multicol}

\usepackage{amsmath}
\usepackage{mathtools}

\title{PropNEAT - Efficient GPU-Compatible Backpropagation over NeuroEvolutionary Augmenting Topology Networks}

\author{ \hspace{1mm}Michael Merry\\
	School of Computer Science\\
	University of Auckland\\
	Symonds St. \\
        Auckland 1071, New Zealand \\
	\texttt{m.merry@auckland.ac.nz} \\
	\And
        \hspace{1mm}Patricia Riddle\\
	School of Computer Science\\
	University of Auckland\\
	Symonds St. \\
        Auckland 1071, New Zealand \\
	\texttt{pat@cs.auckland.ac.nz} \\
	\And
        \hspace{1mm}Jim Warren\\
	School of Computer Science\\
	University of Auckland\\
	Symonds St. \\
        Auckland 1071, New Zealand \\
	\texttt{jim@cs.auckland.ac.nz} \\
	\And\\
}

\renewcommand{\shorttitle}{PropNEAT}

\hypersetup{
pdftitle={PropNEAT - Efficient GPU-Compatible Backpropagation over NeuroEvolutionary Augmenting Topology Networks},
pdfauthor={Michael Merry, Patricia Riddle, Jim Warren},
pdfkeywords={NeuroEvolutionary Algorithms, Sparse Neural Networks, Backpropagation, GPU Optimization},
}

\begin{document}
\maketitle
\begin{abstract}
    We introduce PropNEAT, a fast backpropagation implementation of NEAT that uses a bidirectional mapping of the genome graph to a layer-based architecture that preserves the NEAT genomes whilst enabling efficient GPU backpropagation. We test PropNEAT on 58 binary classification datasets from the Penn Machine Learning Benchmarks database, comparing the performance against logistic regression, dense neural networks and random forests, as well as a densely retrained variant of the final PropNEAT model. PropNEAT had the second best overall performance, behind Random Forest, though the difference between the models was not statistically significant apart from between Random Forest in comparison with logistic regression and the PropNEAT retrain models. PropNEAT was substantially faster than a naive backpropagation method, and both were substantially faster and had better performance than the original NEAT implementation. We demonstrate that the per-epoch training time for PropNEAT scales linearly with network depth, and is efficient on GPU implementations for backpropagation. This implementation could be extended to support reinforcement learning or convolutional networks, and is able to find sparser and smaller networks with potential for applications in low-power contexts.
    
\end{abstract}

% keywords can be removed

\section{Introduction}

The NeuroEvolution of Augmenting Topologies (NEAT) algorithm \cite{Stanley2002} is a genetic algorithm for training sparse neural networks that has been used for a range of purposes,
especially as a competitor to reinforcement learning in control systems \cite{Whiteson2006EvolutionaryLearning, Papavasileiou2021ATopologies}. NEAT evolves complex networks by incrementally adding nodes and edges, optimizing weights through genetic techniques. Its
relative simplicity and lack of need for
advanced hardware or GPUs have driven a high level of interest, including demonstrations of AI on YouTube channels when applied to 
computer games \cite{SethBling2015MarI/OGames}. However, NEAT has had limited application in tabular data due to slow convergence and subpar performance compared to alternatives such as dense neural networks and decision trees. The limitation derives from the use of genetic-based weight optimisation. Gradient-descent based methods have been used but have also been limited due to the inherently sequential implementation of the activation of nodes in NEAT which does not permit the efficiencies of GPU-based backpropagation to be applied.

Here, we present the PropNEAT algorithm, first covering a naive implementation of backpropagation before presenting the PropNEAT algorithm itself. We cover the challenges that arise from the genetic algorithm, the details of the construction and mapping of the layer-based representation that allows for efficient linear-algebra operations, and the details of other secondary changes to NEAT including topological change rates that are subsequently required when using this method. We then present the details of the experiments run to evaluate this algorithm, namely the performance comparison against other predictive models, an ablation experiment against the naive and original implementations, and performance characterisation of the training time of the model. We present the full results and analysis of these experiments, discussing the implications and conclusions that we can draw.

\section{Background}

NEAT \cite{Stanley2002} creates sparse neural networks by adding nodes and connections starting from a minimally connected network where the inputs are connected to the outputs. After initiation, the algorithm follows these steps:
\begin{enumerate}
\item Evaluate the performance of all models,
\item Speciate based on structural similarities using innovation numbers (a unique identifier for each gene),
\item Eliminate the weakest individuals in each species
\item Reproduce with crossover and mutation of weights and topology to replace the eliminated individuals.
\end{enumerate}

NEAT’s key innovation is the use of innovation numbers which are incremental numbers as identifiers of genes to track network topology. This simplifies the alignment of the network graphs for crossover without requiring subgraph analysis. This mechanism enables evolution of both weights and topology using crossover, mutation and speciation. Competition is done within species based on structural and weight differences, configurable with hyperparameters.

The resulting graph model from NEAT networks does not have the same layer-based structure used in other neural networks. Although the per-node calculations are equivalent, they are done sequentially through the graph on a per-node basis, rather than using matrix or tensor operations through layers.

Dense neural networks have had multiple speed improvements since their invention. Raina et al. in 2009 \cite{Raina2009Large-scaleProcessors} demonstrated a 70-times faster implementation of backpropagation over artificial neural networks (ANNs) by using GPUs. They demonstrated techniques for parallelisation for deep networks that allowed the memory advantages and parallelisation of GPUs to be applied to the backpropagation of neural networks. This allows linear algebra optimisations such as from Volkov and Demmel \cite{Volkov2008BenchmarkingAlgebra} over NVidia GPUs to improve the performance of neural network training. Le et al. \cite{Le2011OnLearning} demonstrated how this work, when applied to stochastic gradient descent (SGD), was able to achieve state-of-the-art results over MNIST.

A limitation of the original NEAT implementation has been its
use of the genetic algorithm to update weights, with this being either slow or not
very effective in converging to an optimal solution. Several researchers have added
backpropagation to resolve this problem. Chen and Alahakoon added backpropagation for classification tasks \cite{Chen2006NeuroEvolutionClassification}. They note 
the combined benefits of the breadth of the exploration space from speciation and
the genetic algorithm, with the weight refinement of the backpropagation. They
used a limited amount of backpropagation in their search due to the computational
complexity of the process. They only added in backpropagation in a subset
of generations, and only over a subset of data. They saw a substantial improvement over NEAT.

Whiteson and Stone added a backpropagation implementation to NEAT to support reinforcement learning \cite{Whiteson2006EvolutionaryLearning}. They do not provide details of their backpropagation implementation. They note the improvement in performance over NEAT, and the improved speed in convergence due to the improved weight optimisation process.

Desell added backpropagation to NEAT to allow convolutional neural networks (CNNs) to be created using NEAT \cite{Desell2017DevelopingHyperparameters}. In this work they showed predictive performance comparable with human-designed CNNs. In this work, they describe their backpropagation methodology in section VII.b.
The implementation described is a simple approach which first computes the order of nodes 
in the feed-forward direction and backpropagates the the error in the reverse order. This is what we would describe as a naive implementation. They note the substantial computation requirements for backpropagation over CNNs, and tackled 
this primarily by making adjustments to allow parallel and asynchronous computation of models on multiple machines, using 5,500 in their experiment. This parallel computation was done at a per-genome level, and did not optimise the training at a weights level.

Public and open-source NEAT implementations such as neat-python \cite{McIntyre2019NEATPython} have identified and implemented parallelism of the node computations. However, this has been used to ensure the correct order of activation of nodes and potential CPU threading optimisations. It has not been used to implement tensor-based computations of the nodes. To the best of our knowledge, there is no public implementation that brings tensor-based GPU improvements to NEAT.

\section{PropNEAT Algorithm}

The PropNEAT algorithm efficiently applies backpropagation
on GPUs to train the weights of NEAT-created neural networks. It maps NEAT's graph topology to a layer-based structure allowing a bi-directional mapping of nodes and weights to tensors. This mapping handles skip layers and unreachable nodes, which ultimately permits backpropagation using standard tensor libraries with their GPU optimisations. We will describe the naive approach to backpropagation over NEAT, its limitations, and the PropNEAT algorithm.

\subsection{Naive approach}

Typical implementations of NEAT implement a per-node approach
to calculate the forward pass of the network. The order of operations
can be optimised by adding a graph-analysis step first to ensure all 
node-level dependencies are met, ensuring that all inputs of a node have
been calculated before calculating the result at that node, though other
approaches are also possible. This approach can be directly replicated
in PyTorch or similar, with the computation simplified as 
\begin{equation}
\text{out}[node] = \text{activation}(bias + \sum(\text{inputs}))
\label{eq:node-activation}
\end{equation}
for each node in order. This is the implementation by Desell et al. \cite{Desell2017DevelopingHyperparameters}

This approach takes no advantage of any GPU linear-algebra optimisations,
and instead requires the backpropagation algorithms to differentiate 
and propagate errors through every single node, rather than through each layer.
This does not have any impact on the final result - the computations will be equivalent subject to rounding errors. While it is a sizeable improvement over a genetic search over the weights, and it
is relatively straightforward to implement, it takes no advantage of any of the advantages that GPUs 
provide, which allows effectively for signficantly scaled parallel computation using 
tensor algebra.

\subsection{PropNEAT}

\begin{figure}[!htbp]
  \centering
  \includegraphics[width=0.9\textwidth]{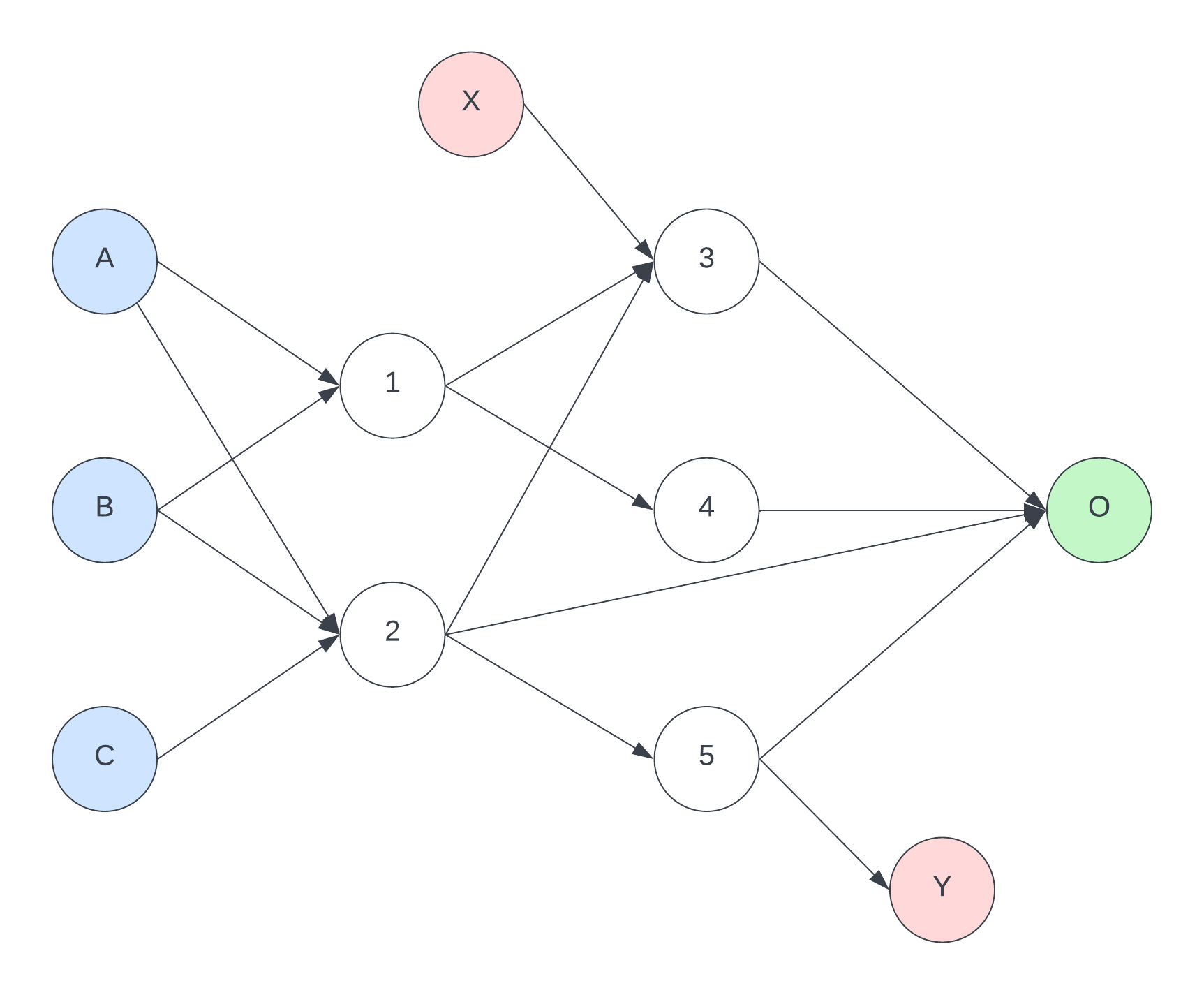}
  \caption{An example network produced by NEAT - Blue: Inputs; Green: Outputs; White: Hidden nodes; Red: Unreachable nodes. This graph exemplifies the major challenges of the naive approach to backpropagation (see 3.3). First, there are unreachable nodes X and Y which either do not connect to the inputs or do not connect to the output through their directed graph. X cannot be reached from the inputs and Y does not connect to the output. Second, there are skip-layer connections such as from 2 to the output. This has two paths 2-3-O and 2-O resulting in a skip-layer effect. If implemented naively this requires 13 operations (one for each connection).}
  \label{fig:neat_problems}
\end{figure}

\subsubsection{Genome-Tensor Mapping}

PropNEAT creates a mapping to a tensor-based structure that is nearly-bijective
and that is compatible with tensor-based computations for GPU compatibility.
By nearly-bijective, we mean that there exists a bijection to the tensor-based
representation, which is augmented by a null-space of zeros within the tensors
that have no computational impact and are not affected by the rest of the training
process. This bijection allows both a genome to be mapped to tensors, and 
the tensor to be mapped to a genome. The only limitations is that gene tracing (via
gene IDs) are lost if the metadata is not maintained.

There are two main challenges in creating such a mapping:
\begin{enumerate}
    \item NEAT creates nodes and connections that can be unreachable
    in the forward or backward direction but may result in advantageous mutations; and
    \item NEAT creates skip connections where the outputs of a single
    node might be at different depths.
\end{enumerate}
These challenges are demonstrated with an example in Fig \ref{fig:neat_problems}.

In addition, we would want the mapping to maintain the bijective properties, map efficiently onto
the minimal tensor representation possible, and be relatively easy to implement.

We solve this by taking several graph traversal and analysis steps before
constructing the mapping:
\begin{enumerate}
    \item Traverse the nodes breadth-first in a forward (input-to-output) direction to compute depth, reachability, and full connectivity
    \item Traverse the nodes breadth-first in a backward (output-to-input) direction
    to compute reachability for back-propagation
    \item Compute the layer structures and connectivity, e.g.,
    skip layers
    \item Map the edges to weights, nodes to their biases, formalising
    the map of genes to tensor indices
\end{enumerate}

\begin{figure}[!htbp]
  \centering
  \includegraphics[width=0.9\textwidth]{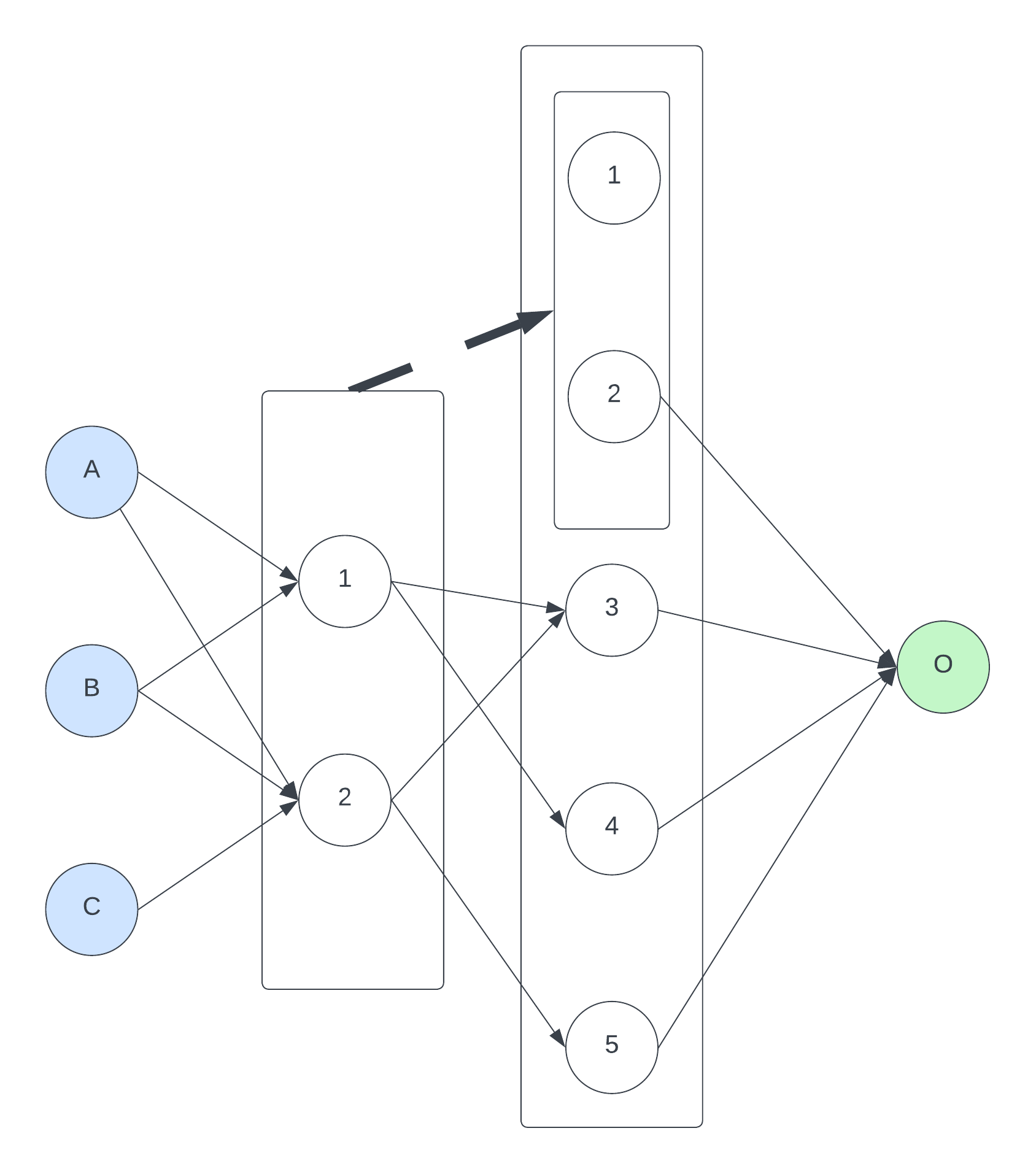}
  \caption{This shows the resulting solution produced by PropNEAT. The graph traversals identify the unreachable nodes and these are removed. The nodes of the same depth from the input are grouped into layers, in this case [1,2] and [3,4,5] as depth 1 and 2 respectively. Where there are skip layers (e.g., 2-O), the outputs of the shallower layer are concatenated to the outputs of subsequent layer and otherwise treated as normal. The subsequent weights layer is then applied across all of these inputs. This provides a consistent layer-based structure that can be mapped to the tensor algebra operations. After the graph-traversal operations, and excluding concatenation as trivial, this requires 3 tensor operations (one for each layer connection).}
  \label{fig:propneat_solutions}
\end{figure}

The resulting structure is demonstrated in Fig \ref{fig:propneat_solutions}.

\subsubsection{Definitions}

\textbf{Depth}: the length of the longest path that
connects an input node to the node. Inputs have $depth=0$.

\textbf{Reachable}: In the forward direction, there exists a path
from at least one input to this node. In the backwards direction,
from at least one output to the node. In this paper, we are only treating single-output models, but this is generalisable to tasks with multiple outputs. This path must be in a consistent direction aligned with either the feed-forward or backpropagation calculations.

\textbf{Unreachable nodes}: nodes that are not reachable in at least one direction. This implies that they either are unable to be affected by the inputs (unreachable forward), unable to affect the outputs (unreachable backward), or unable to affect either.

\textbf{Skip layer}: a layer that has two input layers of different depths.

\textbf{Dense}: a model in which all connections between layers have non-zero weights.

\textbf{Sparse}: a model in which not all connections between layers have non-zero weights, or could be represented by such a model, such as those produced by NEAT.

\textbf{Density}: The proportion of connections between layers which are non-zero.

\subsubsection{Forward pass}
The forward pass does a breadth-first traversal of nodes starting
from the inputs. During this step, it records
the following for each node:
\begin{enumerate}
    \item Depth of the node
    \item List of gene ids of inbound and outbound connections
    \item List of depths of inbound and outbound connections
    \item Is the node an input/output node
    \item Is the node reachable in the forward direction
\end{enumerate}

(1) is used to map this to the correct layer. (2, 3) are used
to compute the layer connectivity. This can also be done during
the backward pass, or partially in both depending on the implementation
of edges. (4) makes input and output edge-case handling easier. (5)
allows unreachable nodes to be discarded for the GPU.

\subsubsection{Backward pass}

The backward pass does a breadth-first traversal of nodes
starting from the outputs. During this step, record for each
node if it is reachable in the backward direction. This allows
the handling of unreachable nodes explicitly. You can also
build or verify depth and connectivity of nodes at this stage as 
part of (2,3) of the forward pass.

\subsubsection{Layer mapping}

Create `k+1` layers, where `k` is the depth of the deepest output node.
The depth of each node defines the layer in which it exists.

For each layer, the following properties are calculated:
\begin{itemize}
    \item The list of nodes in the layer
    \item The order of the nodes, ordered by gene id
    \item Layer type from [input, output, connected]
    \item The output dimension, which is equal
    to the count of nodes in the layer
    \item If this layer has skip inputs, calculated
    by whether any input nodes have $d_{input} < d_{layer} - 1$
    \item If this layer has skip outputs, calculated by
    whether any output nodes have $d_{output} > d_{layer} + 1$
    \item The layer index of any input or output skip layers
\end{itemize}

After this has been computed for all layers, then the input dimension can be computed
for all layers, defined as the sum of output dimensions of all input layers (including
skip connections).

\subsubsection{Instantiating the model}

Instantiating the model is done using the standard approach within a chosen library
but specific handling must be in place for skip connections. Skip connections are handled using concatenation as per DenseNet \cite{Huang2017DenselyNetworks} which most accurately represents the calculations in NEAT. In order
to maintain the bijection, it is important to have a deterministic method to
determine which nodes map to which index within a tensor. In order to accurately 
index the node weight within a tensor, we use the following
conventions to manage the concatenation of skip layers:
\begin{itemize}
    \item The node index is the position of the node, ordered by gene
    id, within the layer
    \item Layers are concatenated in order of layer depth for skip layers.
    \item Layer offset is the sum of output dimensions of the input layers
    with lower layer depth
    \item The resulting index of a node within a tensor is therefore the layer offset +
    the node index    
\end{itemize}

We also rely on the property that a node that has input weights that are all zero
and a bias of zero will have an output of zero under most activation functions.
We will limit our use of activation functions to just those that have this property,
especially focused on ReLU and Sigmoid. In this
case, no error gradient will be associated with this node and all associated weights
will be zero. This is equivalent to this node not existing, or existing but with no 
connections. It is expected that the vast majority of weights will be zeros
due to the sparse networks created by NEAT.

\begin{enumerate}
    \item The input layer is instantiated in the standard way
    \item For each connected layer in order of depth
    \begin{enumerate}
        \item The weight tensor is instantiated according to the input/output 
        dimensions with zero weights. This is a starting point of no connections between the layers.
        \item The weight is overwritten by the weight of each edge, with
        tensor indexing being determined by the conventions above. This is the process of setting the connections between the nodes.
        \item The input tensor is calculated as the concatenation of the output.
        tensors, in increasing order of depth
        \item The bias tensor is instantiated in node index order.
        \item The output tensor is computed in the standard way.
    \end{enumerate}
    \item Any final step with the output layer, with different activation
    functions etc., are implemented.
\end{enumerate}

This method results in a model creation that will have standard forward and back 
functions that is able to be mapped to GPUs, using standard tensor libraries
such as PyTorch.

\subsubsection{Updating genome weights}

After training the model, the genome weights can be updated by reading the 
genome weights from the model according to the conventions described above.

\subsection{Other changes to NEAT}

Replacing the weight optimisation of NEAT with this approach requires a few
changes to the NEAT algorithm.

The weight training happens after genetic cross over, and before evaluation.
This ensures that any topological change is associated with a fine-tuning of 
the model before evaluation, and any improvement can be immediately seen.

By updating the weights via backpropagation, all weight-changing mutations
should be disabled. An exception could be made for reinstantiating all weights,
which can be used to effectively reseed the weights as part of the training but this was not implemented in this experiment.
Otherwise, the weights in the genome should be continued through generations according to
the results of backpropagation. 

Similarly, as topological changes
do not need as many generations for weights to converge, the rate of 
topological mutations can be increased compared to the NEAT parameters. This is because backpropagation is equivalent to multiple generations of genetic improvements of weights. The extent to which the rate of topological mutations can be increased warrants further research. 

The model can be trained for a fixed number of epochs per generation. On the basis of tests during experiments, and a first-principles basis that a model can continue to converge during subsequent generations should it not have occuured, we expect a small number of epochs (e.g., 25) per generation to be chosen. As
weights persist through generations, early stopping methods and similar
are managed with the per-generation evaluation and selection methods. The
improvements of a generation will continue if it results in an overall improvement
according to the GA. The optimisation of this hyperparameter was not explored in this series of experiments.

The balance of topological mutations versus backpropagation time can now
be considered an exploration vs exploitation balance. Increasing topological
mutation rates increases the topological exploration. Increasing the amount
of backpropagation per topological change increases the exploitation within
a topology.

The other mechanisms of NEAT work as per the original implementation \cite{Stanley2002}, including speciation, penalty 
functions and crossover. Adjustments of hyperparameters such as weight vs 
topology penalties for the similarity function for species may need to be
adjusted for optimal performance.

\subsection{Retraining the minimal-covering network}

As NEAT creates a predominantly sparse network, most implied connections in the resulting layer-based model are effectively disabled and the weights matrices will be predominantly zeros as there are a large number of potential connections that have not been made (e.g., between nodes 1 and 5 in Fig \ref{fig:neat_problems}). If one
disregards the requirement to map back to the creating genome, then one
can reseed the weights using any standard strategy (including reseeding
just the zeros and maintaining the weights), and train the fully-connected
network represented on the GPU. As weights that were forced
to be kept as zeros will now become non-zero, the implied topology
will no longer match the source genome and so the bijection will be lost

The methodology
described above will create a set of tensors (network) of minimum size (minimal) 
that can represent the topology of the genome (covering), creating the
minimal covering network. This is the smallest dense network for which
the topology of the genome is a subgraph, and the majority of the weights are zero as there are no genes coding the connection between those nodes.

It is worth noting the similarity to dropout. Dropout takes the tensors of a dense network with all possible connections and masks a random subset to be disconnected, representing them with zeros. This results in a tensor that is a combination of the weights of the activated connections, and zeros for the dropped connections. In PropNEAT, we start with a tensor with no connections, and mask it with the connections that are present in the genome.

\subsection{Recurrence, convolutions, and other variants}

This implementation has been targeted at tabular data, with a single vector of inputs, and a single output. As a first implementation of this technique, the simplest task was chosen for evaluation. As such, this implementation enforces a directed, acyclic graph (DAG).

The methods described here are also applicable to models with recurrence, convolutions, long short-term memory modules, and other node activation functions. These will require further implementation and experimentation.

\section{Methods}

\subsection{Setup}

We benchmarked the PropNEAT algorithm against existing models on open datasets to
characterize it according to:
\begin{itemize}
    \item Predictive performance
    \item Time to train
    \item Model size and complexity
\end{itemize}

The primary hypothesis is that PropNEAT has equivalent predictive performance to other models on
binary classification problems over tabular datasets. This was chosen as a good first demonstration of the algorithm. Rich datasets, including images, video and time series data were excluded on the basis that further changes to NEAT would be required to allow for either recurrence, long-short-term memory, convolutions, and similar which would be necessary for appropriate benchmarking.

The secondary hypotheses are that:
\begin{enumerate}
    \item For the same compute time, PropNEAT has better performance than NEAT
    \item For the same bounding model size, PropNEAT has equivalent performance to
    a fully connected dense neural network
\end{enumerate}

PropNEAT is compared against:
\begin{itemize}
    \item logistic regression (LR),
    \item random forests (RF), and
    \item dense neural networks (NN).
\end{itemize}

The datasets where chosen from the Penn Machine Learning Benchmarking datast (PMLB)
\cite{Olson2017PMLB:Comparison}. The criteria for selection were tabular datasets with a binary-classification target. Image datasets, time series, and other rich datasets were excluded, as were multi-class classification datasets. Each model will be trained
and evaluated on the following datasets:
\begin{multicols}{3}
\begin{enumerate}
    \item adult
    \item appendicitis
    \item australian
    \item backache
    \item biomed
    \item breast
    \item breast\_cancer
    \item breast\_cancer\_wisconsin
    \item breast\_w
    \item buggyCrx
    \item bupa
    \item chess
    \item churn
    \item clean1
    \item clean2
    \item cleve
    \item coil2000
    \item colic
    \item corral
    \item credit\_a
    \item credit\_g
    \item crx
    \item dis
    \item flare
    \item german
    \item glass2
    \item heart\_c
    \item heart\_h
    \item heart\_statlog
    \item house\_votes\_84
    \item hungarian
    \item hypothyroid
    \item ionosphere
    \item irish
    \item kr\_vs\_kp
    \item magic
    \item mofn\_3\_7\_10
    \item molecular\_biology\_promoters
    \item monk2
    \item monk3
    \item mushroom
    \item phoneme
    \item pima
    \item prnn\_crabs
    \item prnn\_synth
    \item profb
    \item ring
    \item saheart
    \item sonar
    \item spambase
    \item spect
    \item spectf
    \item threeOf9
    \item tic\_tac\_toe
    \item tokyo1
    \item twonorm
    \item vote
    \item wdbc
\end{enumerate}
\end{multicols}

Each dataset was split into train and test sets (70:30 ratio), with the same split given to all models and iterations. Validation occurred within the train set, and scripts did not access the test set prior to predictions. To prepare the data, continuous 
variables were z-normalised, categorical variables were one-hot encoded, and 
ordinal variables were one-hot-encoded. The python library sklearn \cite{scikit-learn} was used
for this process. No additional feature engineering or adjustments were made per dataset or model to ensure direct performance comparison.

All stochastic methods were controlled using a fixed seed that was created and recorded as 
the first step in the experiment. The same seed was used for all models and methods and is
recorded against all results. This is also referred
to as the experiment SHA which was used in the results database to help differentiate results of 
experiments during development and testing. For processes where multiple iterations are done, the seed
was incremented by 1 for each iteration of the process.

\subsection{Model training}

Each model was trained in a separate script, with standard methods to load
the train dataset and record predictions. For the stochastic models, the model
was run multiple times. For models requiring hyperparameter selection, this
was done with a dedicated configuration for hyperparameter search ranges and
selection was done using sklearn RandomizedSearchCV. The test data split
was not accessed within these scripts, as all performance evaluation
is done in the analysis process.

\subsubsection{PropNEAT}
PropNEAT was implemented according to the above process in Python 3.7. This
implementation expands from the NEAT Python library \cite{McIntyre2019NEATPython} and uses PyTorch \cite{Paszke2017AutomaticPyTorch} for the
backpropagation and linear algebra.

Training was done for 500 generations with a population of 500 with 25 epochs per generation.
Activation is ReLU for all hidden layers and sigmoid on the output layer. 
Optimisation is done using Adadelta.

NEAT parameters are 0 mean 1 standard deviation (s.d.) for initialisation weights with the following mutation probabilities:
\begin{itemize}
    \item P(add connection) = 0.6
    \item P(remove connection) = 0.6
    \item P(add node) = 0.5
    \item P(remove node) = 0.5
    \item P(reinitialise weights) = 0.0
\end{itemize}

The probabilities were static across all generations and default values from the pyneat library \cite{Tupper2020PYNEAT} were used for all other methods.

The best performing individual over the validation set was chosen from each experiment run as the final model selection for comparison over the test set.

\subsubsection{PropNEAT - Retrain}
The PropNEAT - Retrain evaluates the minimal-bounding NN architecture
of the best PropNEAT model in each run as described in Section 3.5.
The highest performing individual from a PropNEAT run was chosen, and all the weights in the tensors were reinitialised using Kaiming Normal initialisation \cite{Kaiming2018DelvingKaiming}. The training then continued using the PropNEAT implementation. All backpropagation parameters were the same as the PropNEAT implementation. Early stopping was implemented using the train/validate split from the parent PropNEAT training method.

The reuse of the PropNEAT implementation which does not require dropout meant the retrain model did not have dropout implemented either. This is a limitation which means the model will likely not result in the highest performance theoretically possible for the given architecture.

\subsubsection{Dense neural networks}
The dense neural networks used a standard fully-connected model using
dropout (p=0.25) and early stopping. The hidden layer sizes selected were
[256, 512, 512, 256] with input and outputs determined by the datasets. 
Hidden activation was ReLU and final output activation was sigmoid.

\subsubsection{Regression}
Regression models were trained using the LogisticRegression models from
sklearn

\subsubsection{Random forests}
Random forests were trained using the RandomForestRegressor from sklearn with hyperparameters
chosen using the RandomisedSearchCV module.

\subsection{Ablation}

We also compared the difference between PropNEAT, the naive back-propagation implementation of NEAT, and the original NEAT implementation through an ablation experiment over the Adult dataset. The three algorithms were compared with performance over time to test the following hypotheses:
\begin{enumerate}
    \item Backpropagation achieves better results than the genetic-algorithm based weight optimisation
    \item PropNEAT runs faster than a naive implementation of backpropagation
\end{enumerate}

To test these, we ran each of the three algorithms across the Adult data set, measuring time to run and performance. For PropNEAT and the naive implementation, they had the same configuration as above. For the original NEAT implementation, we gave a number of generations equal to the number of epochs times the number of generations in the other implementation.

The experiment was only run over the Adult data set due to the extended computation time for the original implementation.

\subsection{Performance Characterisation}

We characterised the performance of PropNEAT using big-O notation by measuring the training time per individual per epoch, the size, depth and width of the genome, and then took the average of the population per generation. On a training run across all the above datasets, the training time was measured per individual per generation, and then the average time per epoch was calculated. At each generation, this was measured with both the size, depth and width of the genome, with size being the total number of genes, the depth being the total number of layers including input and output, and width being the max width of the layers of the model. This was measured for all iterations of the training of the model.

\subsection{Analysis}

The evaluation of all models was with the AUC over the test set, followed by the rank score of each model. The analyses were done in R unless otherwise stated.

The primary hypothesis tested was that PropNEAT has different performance to the other models. This was tested first with the Friedman rank sum test \cite{Friedman1937TheSt} implement in the R Stats package \cite{RCoreTeam2017R:Computing} with post-hoc pairwise analysis using the Nemenyi test \cite{Nemenyi1963Distribution-freeComparisons} implemented in the PMCMR Plus library in R \cite{Pohlert2023PMCMRplus:Extended}.

The size and complexity of the PropNEAT models were also analysed. The size was defined as total number of parameters, defined as the number of nodes plus number of connections. The "skippiness" was defined as the average number of layers skipped per connection, which captures both the number and depth of skips. For example, a dense model with no skip layers would have skippiness=0, and if every layer was connected directly, and to one extra layer deep, it would have skippines=0.5. The depth of the models is the total number of layers of the models and the average width is the number of nodes divided by the depth.

Correlations between different skippiness, depth, parameter size and AUC were analysed using Pearson correlation, as were those between size, depth, width and time per epoch. Results with p<0.05 were considered statistically significant.

\section{Results}

\begin{table}[]
\centering
\resizebox{\ifdim\width>\linewidth\linewidth\else\width\fi}{!}{
\begin{tabular}[t]{>{}llllll|lllll}
\toprule
\multicolumn{1}{c}{ } & \multicolumn{5}{c}{Max AUC} & \multicolumn{5}{c}{Mean(SD) AUC} \\
\cmidrule(l{3pt}r{3pt}){2-6} \cmidrule(l{3pt}r{3pt}){7-11}
\textbf{Dataset} & \textbf{PR} & \textbf{PN} & \textbf{RF} & \textbf{LR} & \textbf{NN} & \textbf{PR} & \textbf{PN} & \textbf{RF} & \textbf{LR} & \textbf{NN}\\
\midrule

\textbf{adult} & 0.906 & 0.888 & 0.912 & 0.693 & 0.860 & 0.896(0.012) & 0.881(0.005) & 0.912(0.000) & 0.693(0.000) & 0.853(0.004)\\
\textbf{\cellcolor{gray!10}{appendicitis}} & \cellcolor{gray!10}{0.851} & \cellcolor{gray!10}{0.771} & \cellcolor{gray!10}{0.594} & \cellcolor{gray!10}{0.623} & \cellcolor{gray!10}{0.697} & \cellcolor{gray!10}{0.656(0.118)} & \cellcolor{gray!10}{0.766(0.006)} & \cellcolor{gray!10}{0.594(0.000)} & \cellcolor{gray!10}{0.623(0.000)} & \cellcolor{gray!10}{0.606(0.046)}\\
\textbf{australian} & 0.936 & 0.952 & 0.936 & 0.880 & 0.940 & 0.902(0.022) & 0.945(0.006) & 0.936(0.000) & 0.880(0.000) & 0.936(0.002)\\
\textbf{\cellcolor{gray!10}{backache}} & \cellcolor{gray!10}{0.853} & \cellcolor{gray!10}{0.864} & \cellcolor{gray!10}{0.739} & \cellcolor{gray!10}{0.592} & \cellcolor{gray!10}{0.769} & \cellcolor{gray!10}{0.670(0.103)} & \cellcolor{gray!10}{0.766(0.079)} & \cellcolor{gray!10}{0.739(0.000)} & \cellcolor{gray!10}{0.592(0.000)} & \cellcolor{gray!10}{0.682(0.062)}\\

\textbf{biomed} & 0.967 & 0.967 & 0.932 & 0.869 & 0.968 & 0.910(0.041) & 0.962(0.004) & 0.932(0.000) & 0.869(0.000) & 0.876(0.101)\\
\textbf{\cellcolor{gray!10}{breast}} & \cellcolor{gray!10}{0.998} & \cellcolor{gray!10}{0.997} & \cellcolor{gray!10}{0.996} & \cellcolor{gray!10}{0.953} & \cellcolor{gray!10}{0.995} & \cellcolor{gray!10}{0.990(0.006)} & \cellcolor{gray!10}{0.995(0.001)} & \cellcolor{gray!10}{0.996(0.000)} & \cellcolor{gray!10}{0.953(0.000)} & \cellcolor{gray!10}{0.994(0.000)}\\
\textbf{breast\_cancer} & 0.825 & 0.769 & 0.822 & 0.697 & 0.737 & 0.668(0.055) & 0.755(0.010) & 0.822(0.000) & 0.697(0.000) & 0.653(0.042)\\
\textbf{\cellcolor{gray!10}{breast\_cancer\_wisconsin}} & \cellcolor{gray!10}{0.999} & \cellcolor{gray!10}{0.999} & \cellcolor{gray!10}{0.992} & \cellcolor{gray!10}{0.983} & \cellcolor{gray!10}{0.999} & \cellcolor{gray!10}{0.995(0.002)} & \cellcolor{gray!10}{0.998(0.001)} & \cellcolor{gray!10}{0.992(0.000)} & \cellcolor{gray!10}{0.983(0.000)} & \cellcolor{gray!10}{0.998(0.000)}\\
\textbf{breast\_w} & 0.997 & 0.998 & 0.994 & 0.967 & 0.997 & 0.992(0.005) & 0.997(0.001) & 0.994(0.000) & 0.967(0.000) & 0.996(0.000)\\

\textbf{\cellcolor{gray!10}{buggyCrx}} & \cellcolor{gray!10}{0.930} & \cellcolor{gray!10}{0.941} & \cellcolor{gray!10}{0.939} & \cellcolor{gray!10}{0.882} & \cellcolor{gray!10}{0.926} & \cellcolor{gray!10}{0.900(0.018)} & \cellcolor{gray!10}{0.934(0.004)} & \cellcolor{gray!10}{0.939(0.000)} & \cellcolor{gray!10}{0.882(0.000)} & \cellcolor{gray!10}{0.839(0.067)}\\
\textbf{bupa} & 0.638 & 0.634 & 0.588 & 0.556 & 0.641 & 0.545(0.042) & 0.611(0.015) & 0.588(0.000) & 0.556(0.000) & 0.587(0.041)\\
\textbf{\cellcolor{gray!10}{chess}} & \cellcolor{gray!10}{0.999} & \cellcolor{gray!10}{0.995} & \cellcolor{gray!10}{0.997} & \cellcolor{gray!10}{0.971} & \cellcolor{gray!10}{0.995} & \cellcolor{gray!10}{0.993(0.006)} & \cellcolor{gray!10}{0.992(0.003)} & \cellcolor{gray!10}{0.997(0.000)} & \cellcolor{gray!10}{0.971(0.000)} & \cellcolor{gray!10}{0.862(0.097)}\\
\textbf{churn} & 0.914 & 0.884 & 0.916 & 0.623 & 0.687 & 0.871(0.038) & 0.873(0.009) & 0.916(0.000) & 0.623(0.000) & 0.616(0.054)\\
\textbf{\cellcolor{gray!10}{clean1}} & \cellcolor{gray!10}{0.995} & \cellcolor{gray!10}{0.974} & \cellcolor{gray!10}{1.000} & \cellcolor{gray!10}{1.000} & \cellcolor{gray!10}{0.817} & \cellcolor{gray!10}{0.913(0.075)} & \cellcolor{gray!10}{0.923(0.044)} & \cellcolor{gray!10}{1.000(0.000)} & \cellcolor{gray!10}{1.000(0.000)} & \cellcolor{gray!10}{0.738(0.052)}\\

\textbf{clean2} & 1.000 & 1.000 & 1.000 & 1.000 & 1.000 & 1.000(0.000) & 0.997(0.002) & 1.000(0.000) & 1.000(0.000) & 0.999(0.001)\\
\textbf{\cellcolor{gray!10}{cleve}} & \cellcolor{gray!10}{0.893} & \cellcolor{gray!10}{0.902} & \cellcolor{gray!10}{0.892} & \cellcolor{gray!10}{0.816} & \cellcolor{gray!10}{0.894} & \cellcolor{gray!10}{0.857(0.029)} & \cellcolor{gray!10}{0.892(0.008)} & \cellcolor{gray!10}{0.892(0.000)} & \cellcolor{gray!10}{0.816(0.000)} & \cellcolor{gray!10}{0.876(0.017)}\\
\textbf{coil2000} & 0.693 & 0.709 & 0.741 & 0.505 & 0.550 & 0.618(0.048) & 0.686(0.019) & 0.741(0.000) & 0.505(0.000) & 0.529(0.014)\\
\textbf{\cellcolor{gray!10}{colic}} & \cellcolor{gray!10}{0.883} & \cellcolor{gray!10}{0.895} & \cellcolor{gray!10}{0.862} & \cellcolor{gray!10}{0.858} & \cellcolor{gray!10}{0.885} & \cellcolor{gray!10}{0.831(0.028)} & \cellcolor{gray!10}{0.865(0.016)} & \cellcolor{gray!10}{0.862(0.000)} & \cellcolor{gray!10}{0.858(0.000)} & \cellcolor{gray!10}{0.856(0.034)}\\
\textbf{corral} & 1.000 & 0.995 & 1.000 & 0.877 & 1.000 & 0.994(0.020) & 0.983(0.009) & 1.000(0.000) & 0.877(0.000) & 0.999(0.002)\\

\textbf{\cellcolor{gray!10}{credit\_a}} & \cellcolor{gray!10}{0.926} & \cellcolor{gray!10}{0.919} & \cellcolor{gray!10}{0.931} & \cellcolor{gray!10}{0.843} & \cellcolor{gray!10}{0.878} & \cellcolor{gray!10}{0.882(0.017)} & \cellcolor{gray!10}{0.894(0.014)} & \cellcolor{gray!10}{0.931(0.000)} & \cellcolor{gray!10}{0.843(0.000)} & \cellcolor{gray!10}{0.792(0.045)}\\
\textbf{credit\_g} & 0.761 & 0.774 & 0.769 & 0.627 & 0.610 & 0.659(0.031) & 0.753(0.010) & 0.769(0.000) & 0.627(0.000) & 0.556(0.044)\\
\textbf{\cellcolor{gray!10}{crx}} & \cellcolor{gray!10}{0.905} & \cellcolor{gray!10}{0.903} & \cellcolor{gray!10}{0.923} & \cellcolor{gray!10}{0.845} & \cellcolor{gray!10}{0.898} & \cellcolor{gray!10}{0.871(0.020)} & \cellcolor{gray!10}{0.900(0.002)} & \cellcolor{gray!10}{0.923(0.000)} & \cellcolor{gray!10}{0.845(0.000)} & \cellcolor{gray!10}{0.880(0.019)}\\
\textbf{dis} & 0.802 & 0.822 & 0.939 & 0.500 & 0.710 & 0.630(0.098) & 0.782(0.033) & 0.939(0.000) & 0.500(0.000) & 0.691(0.019)\\
\textbf{\cellcolor{gray!10}{flare}} & \cellcolor{gray!10}{0.696} & \cellcolor{gray!10}{0.733} & \cellcolor{gray!10}{0.663} & \cellcolor{gray!10}{0.557} & \cellcolor{gray!10}{0.706} & \cellcolor{gray!10}{0.594(0.054)} & \cellcolor{gray!10}{0.707(0.017)} & \cellcolor{gray!10}{0.663(0.000)} & \cellcolor{gray!10}{0.557(0.000)} & \cellcolor{gray!10}{0.693(0.009)}\\

\textbf{german} & 0.765 & 0.785 & 0.789 & 0.649 & 0.723 & 0.688(0.035) & 0.776(0.008) & 0.789(0.000) & 0.649(0.000) & 0.673(0.031)\\
\textbf{\cellcolor{gray!10}{glass2}} & \cellcolor{gray!10}{0.869} & \cellcolor{gray!10}{0.753} & \cellcolor{gray!10}{0.844} & \cellcolor{gray!10}{0.643} & \cellcolor{gray!10}{0.800} & \cellcolor{gray!10}{0.706(0.100)} & \cellcolor{gray!10}{0.695(0.057)} & \cellcolor{gray!10}{0.844(0.000)} & \cellcolor{gray!10}{0.643(0.000)} & \cellcolor{gray!10}{0.713(0.068)}\\
\textbf{heart\_c} & 0.922 & 0.911 & 0.869 & 0.802 & 0.914 & 0.860(0.033) & 0.894(0.017) & 0.869(0.000) & 0.802(0.000) & 0.891(0.010)\\
\textbf{\cellcolor{gray!10}{heart\_h}} & \cellcolor{gray!10}{0.882} & \cellcolor{gray!10}{0.886} & \cellcolor{gray!10}{0.880} & \cellcolor{gray!10}{0.809} & \cellcolor{gray!10}{0.893} & \cellcolor{gray!10}{0.837(0.026)} & \cellcolor{gray!10}{0.874(0.008)} & \cellcolor{gray!10}{0.880(0.000)} & \cellcolor{gray!10}{0.809(0.000)} & \cellcolor{gray!10}{0.871(0.008)}\\
\textbf{heart\_statlog} & 0.898 & 0.919 & 0.848 & 0.798 & 0.897 & 0.831(0.035) & 0.885(0.027) & 0.848(0.000) & 0.798(0.000) & 0.859(0.071)\\

\textbf{\cellcolor{gray!10}{house\_votes\_84}} & \cellcolor{gray!10}{0.995} & \cellcolor{gray!10}{0.991} & \cellcolor{gray!10}{0.998} & \cellcolor{gray!10}{0.962} & \cellcolor{gray!10}{0.989} & \cellcolor{gray!10}{0.980(0.006)} & \cellcolor{gray!10}{0.982(0.005)} & \cellcolor{gray!10}{0.998(0.000)} & \cellcolor{gray!10}{0.962(0.000)} & \cellcolor{gray!10}{0.986(0.002)}\\
\textbf{hungarian} & 0.798 & 0.799 & 0.842 & 0.732 & 0.801 & 0.742(0.029) & 0.788(0.009) & 0.842(0.000) & 0.732(0.000) & 0.781(0.017)\\
\textbf{\cellcolor{gray!10}{hypothyroid}} & \cellcolor{gray!10}{0.945} & \cellcolor{gray!10}{0.946} & \cellcolor{gray!10}{0.963} & \cellcolor{gray!10}{0.597} & \cellcolor{gray!10}{0.738} & \cellcolor{gray!10}{0.852(0.053)} & \cellcolor{gray!10}{0.933(0.012)} & \cellcolor{gray!10}{0.963(0.000)} & \cellcolor{gray!10}{0.597(0.000)} & \cellcolor{gray!10}{0.701(0.020)}\\
\textbf{ionosphere} & 0.972 & 0.958 & 0.984 & 0.839 & 0.998 & 0.892(0.035) & 0.940(0.020) & 0.984(0.000) & 0.839(0.000) & 0.990(0.011)\\
\textbf{\cellcolor{gray!10}{irish}} & \cellcolor{gray!10}{1.000} & \cellcolor{gray!10}{0.986} & \cellcolor{gray!10}{1.000} & \cellcolor{gray!10}{0.780} & \cellcolor{gray!10}{0.997} & \cellcolor{gray!10}{0.964(0.089)} & \cellcolor{gray!10}{0.973(0.013)} & \cellcolor{gray!10}{1.000(0.000)} & \cellcolor{gray!10}{0.780(0.000)} & \cellcolor{gray!10}{0.942(0.025)}\\

\textbf{kr\_vs\_kp} & 0.998 & 0.988 & 0.998 & 0.967 & 0.989 & 0.990(0.007) & 0.985(0.004) & 0.998(0.000) & 0.967(0.000) & 0.866(0.120)\\
\textbf{\cellcolor{gray!10}{magic}} & \cellcolor{gray!10}{0.923} & \cellcolor{gray!10}{0.875} & \cellcolor{gray!10}{0.928} & \cellcolor{gray!10}{0.747} & \cellcolor{gray!10}{0.875} & \cellcolor{gray!10}{0.917(0.003)} & \cellcolor{gray!10}{0.865(0.008)} & \cellcolor{gray!10}{0.928(0.000)} & \cellcolor{gray!10}{0.747(0.000)} & \cellcolor{gray!10}{0.866(0.005)}\\
\textbf{mofn\_3\_7\_10} & 1.000 & 1.000 & 1.000 & 1.000 & 1.000 & 1.000(0.000) & 1.000(0.000) & 1.000(0.000) & 1.000(0.000) & 0.994(0.013)\\
\textbf{\cellcolor{gray!10}{molecular\_biology\_promoters}} & \cellcolor{gray!10}{0.906} & \cellcolor{gray!10}{0.800} & \cellcolor{gray!10}{0.918} & \cellcolor{gray!10}{0.837} & \cellcolor{gray!10}{0.855} & \cellcolor{gray!10}{0.655(0.108)} & \cellcolor{gray!10}{0.695(0.093)} & \cellcolor{gray!10}{0.918(0.000)} & \cellcolor{gray!10}{0.837(0.000)} & \cellcolor{gray!10}{0.660(0.120)}\\
\textbf{monk2} & 0.994 & 0.629 & 0.994 & 0.444 & 0.662 & 0.886(0.047) & 0.556(0.065) & 0.994(0.000) & 0.444(0.000) & 0.602(0.051)\\

\textbf{\cellcolor{gray!10}{monk3}} & \cellcolor{gray!10}{0.991} & \cellcolor{gray!10}{0.961} & \cellcolor{gray!10}{0.986} & \cellcolor{gray!10}{0.764} & \cellcolor{gray!10}{0.824} & \cellcolor{gray!10}{0.980(0.007)} & \cellcolor{gray!10}{0.957(0.004)} & \cellcolor{gray!10}{0.986(0.000)} & \cellcolor{gray!10}{0.764(0.000)} & \cellcolor{gray!10}{0.759(0.042)}\\
\textbf{mushroom} & 1.000 & 0.991 & 1.000 & 0.941 & 1.000 & 0.962(0.057) & 0.987(0.004) & 1.000(0.000) & 0.941(0.000) & 0.997(0.002)\\
\textbf{\cellcolor{gray!10}{phoneme}} & \cellcolor{gray!10}{0.931} & \cellcolor{gray!10}{0.857} & \cellcolor{gray!10}{0.955} & \cellcolor{gray!10}{0.662} & \cellcolor{gray!10}{0.836} & \cellcolor{gray!10}{0.913(0.007)} & \cellcolor{gray!10}{0.844(0.007)} & \cellcolor{gray!10}{0.955(0.000)} & \cellcolor{gray!10}{0.662(0.000)} & \cellcolor{gray!10}{0.832(0.002)}\\
\textbf{pima} & 0.824 & 0.836 & 0.823 & 0.710 & 0.826 & 0.762(0.036) & 0.833(0.003) & 0.823(0.000) & 0.710(0.000) & 0.815(0.015)\\
\textbf{\cellcolor{gray!10}{prnn\_crabs}} & \cellcolor{gray!10}{1.000} & \cellcolor{gray!10}{1.000} & \cellcolor{gray!10}{0.966} & \cellcolor{gray!10}{1.000} & \cellcolor{gray!10}{1.000} & \cellcolor{gray!10}{1.000(0.000)} & \cellcolor{gray!10}{1.000(0.000)} & \cellcolor{gray!10}{0.966(0.000)} & \cellcolor{gray!10}{1.000(0.000)} & \cellcolor{gray!10}{0.817(0.077)}\\

\textbf{prnn\_synth} & 0.958 & 0.968 & 0.941 & 0.870 & 0.966 & 0.896(0.044) & 0.964(0.003) & 0.941(0.000) & 0.870(0.000) & 0.964(0.005)\\
\textbf{\cellcolor{gray!10}{profb}} & \cellcolor{gray!10}{0.674} & \cellcolor{gray!10}{0.696} & \cellcolor{gray!10}{0.657} & \cellcolor{gray!10}{0.597} & \cellcolor{gray!10}{0.669} & \cellcolor{gray!10}{0.582(0.037)} & \cellcolor{gray!10}{0.669(0.023)} & \cellcolor{gray!10}{0.657(0.000)} & \cellcolor{gray!10}{0.597(0.000)} & \cellcolor{gray!10}{0.630(0.024)}\\
\textbf{ring} & 0.987 & 0.904 & 0.985 & 0.763 & 0.995 & 0.980(0.005) & 0.893(0.012) & 0.985(0.000) & 0.763(0.000) & 0.992(0.001)\\
\textbf{\cellcolor{gray!10}{saheart}} & \cellcolor{gray!10}{0.765} & \cellcolor{gray!10}{0.799} & \cellcolor{gray!10}{0.718} & \cellcolor{gray!10}{0.688} & \cellcolor{gray!10}{0.774} & \cellcolor{gray!10}{0.682(0.043)} & \cellcolor{gray!10}{0.782(0.014)} & \cellcolor{gray!10}{0.718(0.000)} & \cellcolor{gray!10}{0.688(0.000)} & \cellcolor{gray!10}{0.748(0.018)}\\
\textbf{sonar} & 0.887 & 0.791 & 0.862 & 0.713 & 0.737 & 0.765(0.059) & 0.747(0.039) & 0.862(0.000) & 0.713(0.000) & 0.683(0.049)\\

\textbf{\cellcolor{gray!10}{spambase}} & \cellcolor{gray!10}{0.976} & \cellcolor{gray!10}{0.978} & \cellcolor{gray!10}{0.981} & \cellcolor{gray!10}{0.916} & \cellcolor{gray!10}{0.970} & \cellcolor{gray!10}{0.965(0.005)} & \cellcolor{gray!10}{0.975(0.003)} & \cellcolor{gray!10}{0.981(0.000)} & \cellcolor{gray!10}{0.916(0.000)} & \cellcolor{gray!10}{0.954(0.041)}\\
\textbf{spect} & 0.807 & 0.804 & 0.744 & 0.583 & 0.852 & 0.701(0.051) & 0.787(0.018) & 0.744(0.000) & 0.583(0.000) & 0.843(0.010)\\
\textbf{\cellcolor{gray!10}{spectf}} & \cellcolor{gray!10}{0.919} & \cellcolor{gray!10}{0.860} & \cellcolor{gray!10}{0.921} & \cellcolor{gray!10}{0.725} & \cellcolor{gray!10}{0.867} & \cellcolor{gray!10}{0.829(0.032)} & \cellcolor{gray!10}{0.832(0.020)} & \cellcolor{gray!10}{0.921(0.000)} & \cellcolor{gray!10}{0.725(0.000)} & \cellcolor{gray!10}{0.786(0.044)}\\
\textbf{threeOf9} & 0.995 & 0.931 & 1.000 & 0.804 & 0.866 & 0.971(0.019) & 0.921(0.012) & 1.000(0.000) & 0.804(0.000) & 0.743(0.100)\\
\textbf{\cellcolor{gray!10}{tic\_tac\_toe}} & \cellcolor{gray!10}{0.975} & \cellcolor{gray!10}{0.763} & \cellcolor{gray!10}{0.999} & \cellcolor{gray!10}{0.603} & \cellcolor{gray!10}{0.758} & \cellcolor{gray!10}{0.881(0.036)} & \cellcolor{gray!10}{0.733(0.025)} & \cellcolor{gray!10}{0.999(0.000)} & \cellcolor{gray!10}{0.603(0.000)} & \cellcolor{gray!10}{0.707(0.039)}\\

\textbf{tokyo1} & 0.970 & 0.980 & 0.977 & 0.904 & 0.972 & 0.940(0.028) & 0.974(0.005) & 0.977(0.000) & 0.904(0.000) & 0.968(0.003)\\
\textbf{\cellcolor{gray!10}{twonorm}} & \cellcolor{gray!10}{0.997} & \cellcolor{gray!10}{0.998} & \cellcolor{gray!10}{0.996} & \cellcolor{gray!10}{0.977} & \cellcolor{gray!10}{0.998} & \cellcolor{gray!10}{0.996(0.001)} & \cellcolor{gray!10}{0.998(0.000)} & \cellcolor{gray!10}{0.996(0.000)} & \cellcolor{gray!10}{0.977(0.000)} & \cellcolor{gray!10}{0.998(0.000)}\\
\textbf{vote} & 0.998 & 0.997 & 0.999 & 0.963 & 0.994 & 0.987(0.006) & 0.994(0.002) & 0.999(0.000) & 0.963(0.000) & 0.992(0.002)\\
\textbf{\cellcolor{gray!10}{wdbc}} & \cellcolor{gray!10}{0.999} & \cellcolor{gray!10}{0.999} & \cellcolor{gray!10}{0.993} & \cellcolor{gray!10}{0.979} & \cellcolor{gray!10}{0.999} & \cellcolor{gray!10}{0.995(0.002)} & \cellcolor{gray!10}{0.998(0.001)} & \cellcolor{gray!10}{0.993(0.000)} & \cellcolor{gray!10}{0.979(0.000)} & \cellcolor{gray!10}{0.998(0.000)}\\
\bottomrule
\end{tabular}}

\caption{Model results by dataset. PR: PropNEAT Retrain; PN: PropNEAT; RF: Random Forest; LR: Logistic Regression; NN: Neural Network}
\label{table:model_results_by_dataset}
\end{table}

Model performances varied significantly (Friedman chi-squared = 147.58; p=8.929e-16). Table \ref{table:model_results_by_dataset} shows the results of all models on all datasets. Table \ref{table:model_performance} reports the summarised mean, median and S.D. of AUC and mean and S.D. of ranks across all datasets. By order of mean rank Random Forest was the highest performing, followed by PropNEAT, Neural Network, Logistic Regression and PropNEAT Retrain.  Table \ref{table:pairwise_performance} presents pair-wise comparisons. Only Random Forests and PropNEAT retrain showed a significant difference (p=0.037). Six datasets (chess, breast, breast\_w, breast\_cancer\_wisconsin, cleve, and clean2) showed AUC spreads <0.01 indicating equivalent performance across all models. Table \ref{table:pairwise_performance_with_exclusions} presents the pair-wise comparisons of datasets with these six datasets excluded. These did not change the order of performance, but yielded significant differences between Random Forests and Logistic Regression (p=0.031) and between Random Forests and PropNEAT retrain (p=0.021).

\begin{table}[ht]
\centering
\begin{tabular}{l|ccccc}
\hline
\textbf{Prediction Method} & \textbf{Mean AUC} & \textbf{Median AUC} & \textbf{S.D. of AUC} & \textbf{Mean Rank} & \textbf{S.D. of Rank} \\
\hline
Logistic Regression & 0.832 & 0.839 & 0.106 & 3.317 & 1.342 \\
Neural Network    & 0.811 & 0.826 & 0.132 & 3.200 & 1.448 \\
PropNEAT          & 0.842 & 0.863 & 0.101 & 2.983 & 0.951 \\
PropNEAT Retrain & 0.830 & 0.880 & 0.132 & 3.367 & 1.420 \\
Random Forest     & 0.866 & 0.920 & 0.129 & 2.133 & 1.548 \\
\hline
\end{tabular}

\caption{Model performances}
\label{table:model_performance}
\end{table}

\begin{table}[ht]
\centering
\begin{tabular}{l|cccc}
\hline
\textbf{Prediction Method} & \textbf{Logistic Regression} & \textbf{Neural Network} & \textbf{PropNEAT} & \textbf{PropNEAT Retrain} \\
\hline
Neural Network    & 0.981 & -     & -     & -     \\
PropNEAT          & 0.899 & 0.997 & -     & -     \\
PropNEAT Retrain & 1.000 & 0.946 & 0.817 & -     \\
Random Forest     & 0.062 & 0.226 & 0.402 & \textbf{0.037} \\
\hline
\end{tabular}

\caption{p-values of pair-wise performance comparisons of ranks}
\label{table:pairwise_performance}
\end{table}

\begin{table}[ht]
\centering
\begin{tabular}{l|cccc}
\hline
\textbf{Prediction Method} & \textbf{Logistic Regression} & \textbf{Neural Network} & \textbf{PropNEAT} & \textbf{PropNEAT Retrain} \\
\hline
Neural Network    & 0.999 & -     & -     & -     \\
PropNEAT          & 0.926 & 0.984 & -     & -     \\
PropNEAT Retrain & 1.000 & 0.994 & 0.882 & -     \\
Random Forest     & \textbf{0.031} & 0.068 & 0.228 & \textbf{0.021} \\
\hline
\end{tabular}
\caption{p-values of pair-wise performance comparisons of ranks (with dataset exclusions)}
\label{table:pairwise_performance_with_exclusions}
\end{table}

\begin{figure}[!htbp]
  \centering
  \includegraphics[width=0.9\textwidth]{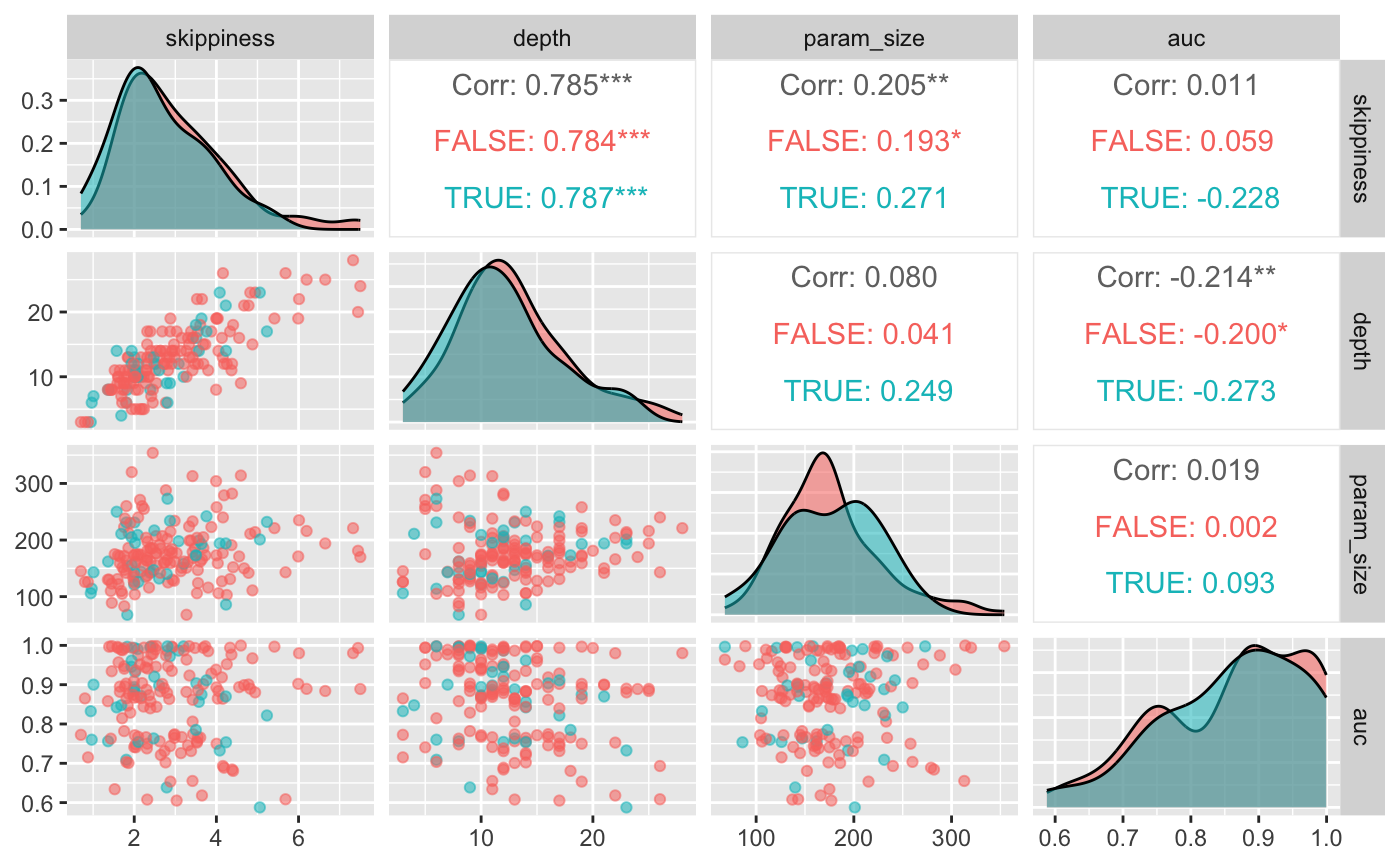}
  \caption{Scatter, correlation and histogram plots for PropNEAT Model complexity over all iterations for all datasets. "True" models are the highest-performing on validation data, used for final analysis. "False" models are other candidates with lower validation performance. Significance is shown with * indicating p<0.05, ** indicating p<0.01, *** indicating p<0.001}
  \label{fig:complexity_corrplot}
\end{figure}

Table \ref{table:complexity_statistics} shows the summary statistics of the complexity of the highest-performing PropNEAT model on each dataset, with depth, number of parameters, and skippines of the model. Table \ref{table:complexity_statistics_others} shows the same for the trained models that weren't chosen. Figure \ref{fig:complexity_corrplot} presents correlation plots of skippines, depth, parameter size and AUC against each other. Skippiness is found to be strongly correlated with depth (p<0.001). Parameter size was generally found to be correlated skippiness and depth was negatively correlated with AUC, though these were not found to be significant for only the highest performing models. 

% Table \ref{table:correlation} shows the correlations between the number of iterations, skippines, depth parameter size and the AUC.

\begin{figure}[!htbp]
  \centering
  \includegraphics[width=0.9\textwidth]{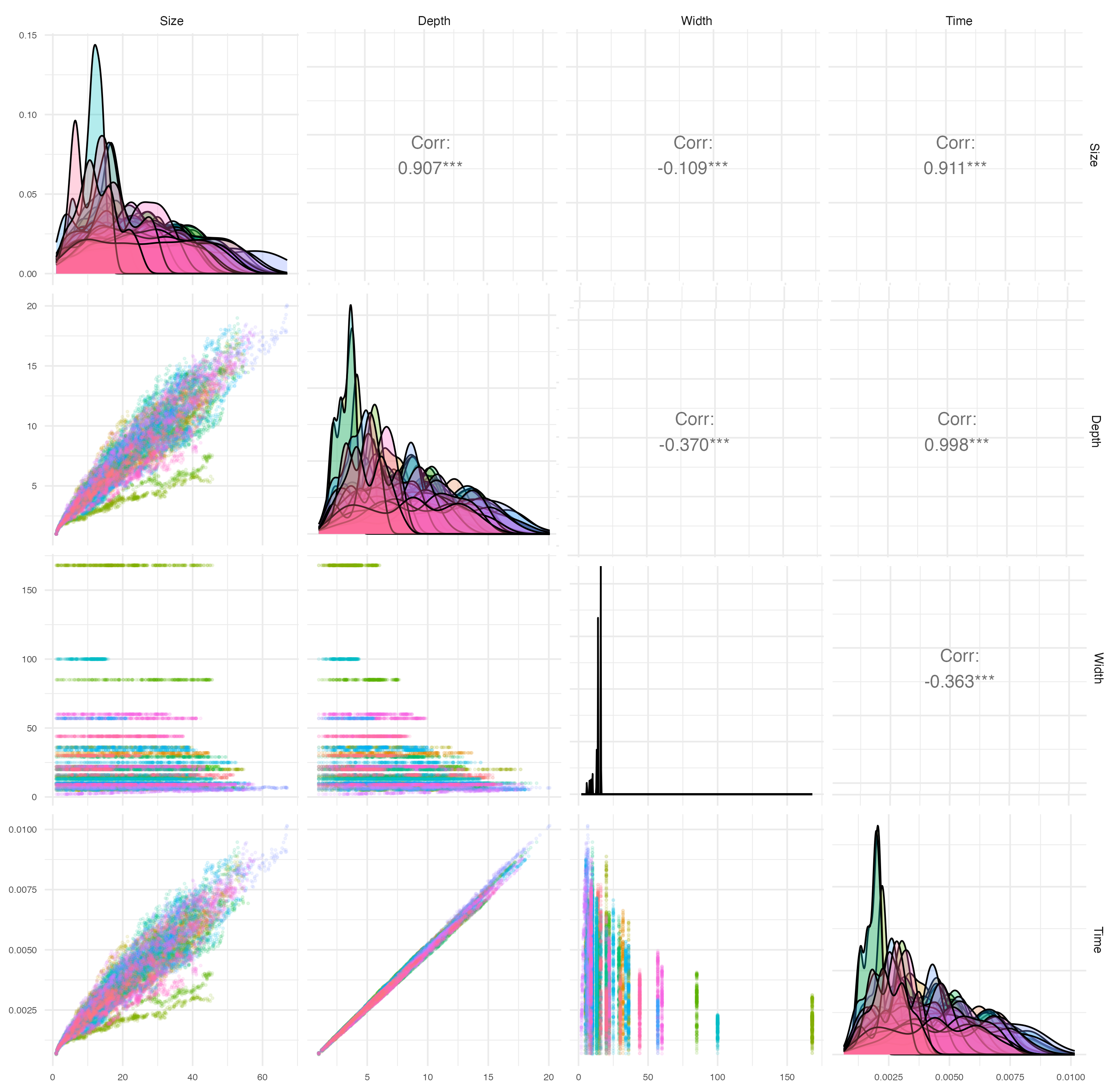}
  \caption{Correlation plot of average time per epoch (s), width (n nodes), depth (n layers) and size (n nodes). Colours are datasets. Significance is shown with * indicating p<0.05, ** indicating p<0.01, *** indicating p<0.001}
  \label{fig:explaneat_performance_corrplots}
\end{figure}

Fig \ref{fig:explaneat_performance_corrplots} shows the relationships, and correlations between size, depth, width, and time per epoch when training PropNEAT. Depth and time per epoch were correlated with a Pearson coefficient of 0.998, (p<0.001), as were size of the genome and time with a Pearson coefficient of 0.911 (p<0.001). The relationship between Depth and Time demonstrates the big-O factor for PropNEAT which is $O(d)$ for the depth, d, in number of layers. This is highly correlated with the size of the genome, but would become less so as the size of the genome increases. We expect that this would be approximately $O(log(n))$ for the number of genes, n, in the genome. 
 
\begin{table}[!h]
\centering
\begin{tabular}{l|ccccc}
\hline
Measurement Type & Min & Mean & Median & S.D. & Max \\
\hline
Depth       & 3.000 & 11.944 & 11.500 & 4.951 & 23.000 \\
Parameter Size & 68.000 & 176.167 & 183.000 & 48.689 & 273.000 \\
Skippiness  & 0.942 & 2.637 & 2.432 & 1.093 & 5.230 \\
\hline
\end{tabular}
\caption{Complexity of the PropNEAT models that were the best in each population for the best performing iteration against the validation dataset.}
\label{table:complexity_statistics}
\end{table}

\begin{table}[!h]
\centering
\begin{tabular}{lrrrrr}
\hline
Measurement Type & Min & Mean & Median & S.D. & Max \\ 
\hline
Depth & 3.000 & 12.908 & 12.000 & 5.047 & 28.000 \\
Parameter Size & 68.000 & 176.415 & 171.000 & 50.636 & 354.000 \\
Skippiness & 0.701 & 2.990 & 2.732 & 1.321 & 7.504 \\
\hline
\end{tabular}
\caption{Complexity of PropNEAT models for not-selected PropNEAT models.}
\label{table:complexity_statistics_others}
\end{table}

% \begin{figure}[!htbp]
%   \centering
%   \includegraphics[width=0.9\textwidth]{average_width_histogram.png}
%   \caption{Histogram of average width of models}
%   \label{fig:complexity_corrplot}
% \end{figure}

% \begin{table}[ht]
% \centering
% \begin{tabular}{l|ccccc}
% \hline
%            & Iterations & Skippiness & Depth  & Parameter Size & AUC    \\
% \hline
% Iterations  & ---       & 0.8877     & 0.7298 & 0.7701      & 0.8987 \\
% Skippiness &           & ---        & 0.0000 & 0.0062      & 0.8876 \\
% Depth      &           &            & ---    & 0.2888      & 0.0042 \\
% Parameter Size &          &            &        & ---         & 0.7992 \\
% AUC        &           &            &        &             & ---    \\
% \hline
% \end{tabular}
% \caption{Table of correlation coefficients between different }
% \label{table:correlation}
% \end{table}

% \begin{table}[ht]
% \centering
% \begin{tabular}{l|ccccc}
% \hline
%            & iteration & skippiness & depth  & param\_size & auc    \\
% \hline
% iteration  & ---       & 0.2385     & 0.4734 & 0.0893      & 0.1857 \\
% skippiness & 0.5745    & ---        & 0.0000 & 0.1092      & 0.1821 \\
% depth      & 0.3665    & 0.0000     & ---    & 0.1435      & 0.1066 \\
% param\_size & 0.6451    & 0.0214     & 0.6320 & ---         & 0.5909 \\
% auc        & 0.6272    & 0.4858     & 0.0169 & 0.9817      & ---    \\
% \hline
% \end{tabular}
% \caption{Combined Correlation Table}
% \label{table:combined_correlation}
% \end{table}

Table \ref{tab:ablation_results} presents the AUCs and time to train for the three models over the adult dataset.

\begin{table}[h!]
    \centering
    \begin{tabular}{lcr}
        \toprule
        \textbf{Model} & \textbf{AUC}  & \textbf{time} \\
        \midrule

        PropNEAT & 0.888  & 1h33m35\\
        Naive Backpropagation & 0.884 & 6h11m54 \\
        Original NEAT & 0.614  & 23h36m20\\
        \bottomrule
    \end{tabular}
    \caption{Ablation Results over adult}
    \label{tab:ablation_results}
\end{table}

\section{Discussion}

PropNEAT was shown to have predictive performance between Logistic Regression and Random Forests, and comparable to Neural Networks, on these datasets. This is to be expected, as the NEAT models are a subset of all possible Neural Networks, which are generally expected to be better than logistic regression without substantial additional feature generation. It is likely that increased performance is possible for most of the models, especially for PropNEAT. The PropNEAT models were trained using a fixed number of generations and epochs due to challenges in specifying other appropriate stopping criteria. It is likely that the models have not "fully converged", though this term is perhaps not entirely appropriate for describing the genetic algorithm finding the optimal topology. It is possible that further performance can be attained with better stopping criteria, such as number of generations without a topological innovation driving improvements. This experiment shows the relative performance with as little human intervention as possible; it is expected that all models could be improved with additional effort from the modellers.

Different datasets had different spreads of results between models. There are some datasets where the logistic regression was not well differentiated from other models, such as twonorm, or breast\_cancer\_wisconsin. There are other datasets where there is a significant different between the results, such as coil2000 or magic. Although there was no significant difference between the results of the models, there are clearly some datasets where there is signal in the data that cannot be modelled with logistic regression but where the machine-learning models are able to do so.

There is not a clear advantage to retraining dense versions of the network topologies from PropNEAT, and possibly even a disadvantage. This is somewhat counter-intuitive as the retrained model is denser with a greater number of parameters. The retraining method reused the method from PropNEAT and so didn't include dropout or early stopping - it is likely that part of the drop in performance is due to overfitting. It is worth further work to see if these retrained models can be improved with a dedicated implementation that included these improvements to manage the overfitting problem.

The PropNEAT models were not very dense, with an average of 176 parameters. The spread of parameter size from 68 to 273 indicates that the exploration and exploitation of topologies was working, at least to some extent. There was enough flexibility for the model to expand to 273 parameters, but with enough pressure that if the complexity was not needed, the model held to only 68 parameters. This gives validity to NEAT's ability to create parsimonious models. This could likewise be optimised, for example with an increasing penalty for complexity as generations increase allowing for early exploration but increasingly selecting for more parsimonious models. 

That there was no particular correlation between the AUC and topology of the models is to be expected. As there is no particular reason to expect that a dataset with a maximum possible AUC for models has a specific complexity or topology (rather, it is a function of the total explainable variance of the target variable from the data provided). This means that the model is likely explaining as much of the variance as is possible, without excess. This is further evidence for the parsimony of the PropNEAT models.

The combination of topological mutation through the genetic algorithm and the optimisation with the backpropogation has a clear parallel to exploration and exploitation tradeoffs. Hyperparameter tuning can likely substantially improve performance by balancing the mutations with generation length, and optimal backpropagation strategies. There is a wide range of possibilities for the backpropagation strategies beyond current methods as a topology can persist over multiple generations. Concepts such as momentum might need to be reconsidered in this context, and hyperparameters such as mutation likelihood may have possibilities for longitudinal strategies allowing for change in exploration-exploitation balances over time, decreasing mutation and increasing backpropagation over the course of training. There may also be other considerations as the overhead of transferring the models on and off the GPU may become an important factor in training speed, though code optimisation and memory management may also help mitigate this.

The similar performance of PropNEAT compared to the retrained version, and the Neural Networks, despite orders of magnitude fewer parameters has some potential advantages. Smaller, sparse networks may have value in low-energy or otherwise resource-constrained devices. Further, as the complexity of the networks is lower, the models are potentially more explainable. There is the possibility for graph analysis of the network topologies. The specific structure of connections and weights may yield clearer relationships within the network than is possible to identify with dense networks, but there are potentially computational challenges with graph and sub-graph analysis that would need to be tackled.

The ablation experiment demonstrates that there is no real difference in performance between a naive implementation and PropNEAT, but both are substantially better than the original NEAT implementation. This is to be expected - there are computation efficiencies between the naive implementation and PropNEAT, but the two methods are equivalent up to numerical methods at computation time. In comparison, backpropagation versus genetic optimisation of weights has an appreciable difference. There is also a large improvement in speed with PropNEAT, and this makes it a viable option.

The performance characterisation clearly demonstrates the scaling factor of the algorithm being the depth of the genome ($O(d)$), rather than the size of the genome ($O(n)$). Depth should scale sublinearly compared to the number of genes. The relationship between the depth and size of genome in NEAT is a function of the probabilities of adding a new node (which has a chance to increase the overall depth), and the probability of adding a new connection (which increases the connection density and genome size). In the extreme case where all mutations were new nodes, there would be little performance increase. In the average case, we might expect a square relationship between genes and depth, but this will also be influenced by what the optimal network shape is. We also see that datasets with a greater number of inputs show a greater improvement in the speed as they producer wider networks. This indicates there may be especially good use cases in problems such as genetic data predictions where the input size is very large.

This experiment is limited in scope, focusing only on tabular datasets. However, the backpropagation method demonstrated here could be extended to support convolutional and recurrent neural networks with appropriate adaptations to the base NEAT algorithm. By adjusting what the representation of a node is, one can expand this to be larger structures, with a single node potentially representing an LSTM module, convolution, or similar. It is likely possible to extend this method to allow parallel training of the individuals within a population on a single GPU, which could provide substantial speed improvements. It is not clear how well this will scale to more complex architectures and this is left for future research.

\section{Conclusion}

PropNEAT demonstrates a viable method for using backpropagation on NEAT-topology neural networks whilst making use of the advantages of GPUs on tabular data. It has predictive performance similar to dense neural networks, but with smaller networks. There are large speed improvements over a naive implementation of backpropagation for NEAT-based models, with training time being determined by the depth of the networks. Analysis of these structures may be of interest for explainability of specific networks and understanding the importance of skip and sparse structures. Further work is needed to extend this to convolutional and recurrent network structures.

% \subsection{Citations}
% Citations use \verb+natbib+. The documentation may be found at
% \begin{center}
% 	\url{http://mirrors.ctan.org/macros/latex/contrib/natbib/natnotes.pdf}
% \end{center}

% Here is an example usage of the two main commands (\verb+citet+ and \verb+citep+): Some people thought a thing \citep{kour2014real, hadash2018estimate} but other people thought something else \citep{kour2014fast}. Many people have speculated that if we knew exactly why \citet{kour2014fast} thought this\dots

% \subsection{Figures}
% \lipsum[10]
% See Figure \ref{fig:fig1}. Here is how you add footnotes. \footnote{Sample of the first footnote.}
% \lipsum[11]

% \begin{figure}
% 	\centering
% 	\fbox{\rule[-.5cm]{4cm}{4cm} \rule[-.5cm]{4cm}{0cm}}
% 	\caption{Sample figure caption.}
% 	\label{fig:fig1}
% \end{figure}

% \subsection{Tables}
% See awesome Table~\ref{tab:table}.

% The documentation for \verb+booktabs+ (`Publication quality tables in LaTeX') is available from:
% \begin{center}
% 	\url{https://www.ctan.org/pkg/booktabs}
% \end{center}

% \begin{table}
% 	\caption{Sample table title}
% 	\centering
% 	\begin{tabular}{lll}
% 		\toprule
% 		\multicolumn{2}{c}{Part}                   \\
% 		\cmidrule(r){1-2}
% 		Name     & Description     & Size ($\mu$m) \\
% 		\midrule
% 		Dendrite & Input terminal  & $\sim$100     \\
% 		Axon     & Output terminal & $\sim$10      \\
% 		Soma     & Cell body       & up to $10^6$  \\
% 		\bottomrule
% 	\end{tabular}
% 	\label{tab:table}
% \end{table}

% \subsection{Lists}
% \begin{itemize}
% 	\item Lorem ipsum dolor sit amet
% 	\item consectetur adipiscing elit.
% 	\item Aliquam dignissim blandit est, in dictum tortor gravida eget. In ac rutrum magna.
% \end{itemize}

% \bibliographystyle{unsrtnat}
\bibliographystyle{unsrt}
\bibliography{references}  %%% Uncomment this line and comment out the ``thebibliography'' section below to use the external .bib file (using bibtex) .





% \end{thebibliography}

\appendix

\section{Worked Example of Figure \ref{fig:propneat_solutions}}

\subsection{Matrix operations}

For clarity, we have repeated Figure \ref{fig:propneat_solutions}

\begin{figure}[!htbp]
  \centering
  \includegraphics[width=0.5\textwidth]{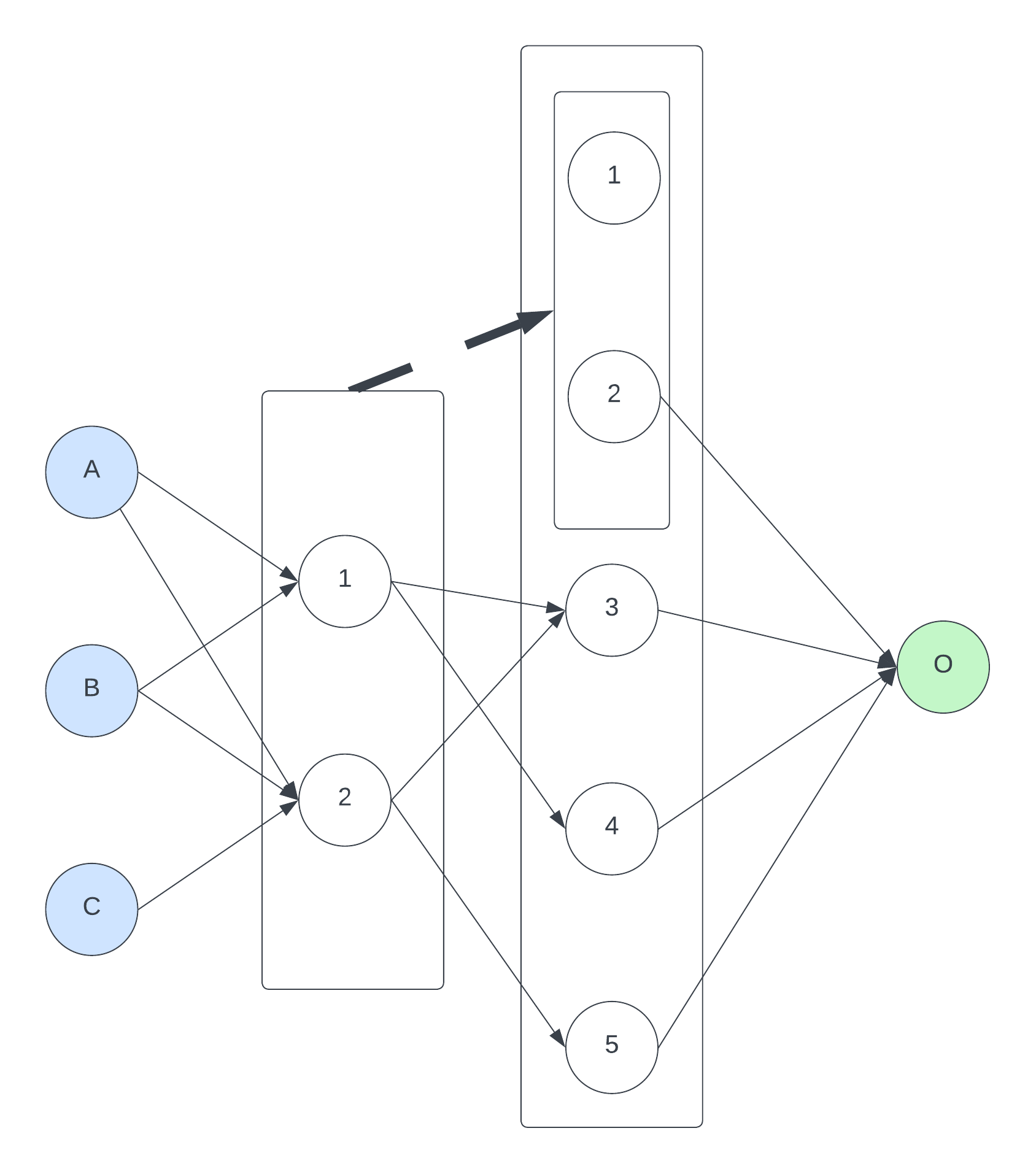}
  \caption{This shows the resulting mapping produced by PropNEAT. The graph traversals identify the unreachable nodes and these are removed. The nodes of the same depth from the input are grouped into layers, in this case [1,2] and [3,4,5] as depth 1 and 2 respectively. Where there are skip layers (e.g., 2-O), the outputs of the shallower layer concatenated to the outputs of subsequent layer and otherwise treated as normal. The subsequent weights layer is then applied across all of these inputs. This provides a consistent layer-based structure that can be mapped to the tensor algebra operations. After the graph-traversal operations, and excluding concatenation as trivial, this requires 3 tensor operations (one for each layer connection).}
  \label{fig:propneat_solutions_repeat}
\end{figure}

We define the operator $\oplus$ to be the concatenation operation between two vectors. We will use $a, b, c$ to be the values of the inputs. We will use $w_{pq}$ to represent the weight that connects node $p$ to node $q$. We will use $h_n$ to be the output of the hidden node $n$. We will use $H_l$ to denote the output vector of the $l$th layer, and $W_{kl}$ to denote the weights between the $k$ and $l$th layers. We define Layer $O$ as the output layer and layer $I$ as the input layer.

The linear algebra operations for the network in Figure \ref{fig:propneat_solutions_repeat} can be stated as:

\begin{equation}
F(H_I W_{I1}) = 
F\left(
\begin{bmatrix}
a \\
b \\
c
\end{bmatrix}
\begin{bmatrix}
w_{a1} & w_{b1} & w_{c1} \\
w_{a2} & w_{b2} & w_{c2} 
\end{bmatrix}\right)
=
F\left(
\begin{bmatrix}
h_1 \\
h_2
\end{bmatrix}
\right)
\coloneq H_1
\end{equation}

\begin{equation}
F(H_1 W_{12}) = 
F\left(
\begin{bmatrix}
h_1 \\
h_2
\end{bmatrix}
\begin{bmatrix}
w_{31} & w_{32} \\
w_{41} & w_{42} \\
w_{51} & w_{52}
\end{bmatrix}
\right)
=
F\left(
\begin{bmatrix}
h_3 \\
h_4 \\
h_5
\end{bmatrix}
\right)
\coloneq  H_2
\end{equation}

\begin{equation}
H_2' = H_1 \oplus H_2 =
\begin{bmatrix}
h_1 \\
h_2 \\
h_3 \\
h_4 \\
h_5
\end{bmatrix}
\end{equation}

\begin{equation}
F\left(H_2'W_{2O}\right)=
F\left(
\begin{bmatrix}
h_1 \\
h_2 \\
h_3 \\
h_4 \\
h_5
\end{bmatrix}
\begin{bmatrix}
w_{1O} & w_{2O} & w_{3O} & w_{4O} & w_{5O}
\end{bmatrix}
\right)
= O
\end{equation}

However, not every connection has $|w_{ij}|>0$. Specifically, there is only a non-zero weight between connected nodes. Replacing the weights between disconnected nodes with zeroes, we get the following:

\begin{equation}
F(H_I W_{I1}) = 
F\left(
\begin{bmatrix}
a \\
b \\
c
\end{bmatrix}
\begin{bmatrix}
w_{a1} & w_{b1} & 0 \\
w_{a2} & w_{b2} & w_{c2} 
\end{bmatrix}\right)
=
F\left(
\begin{bmatrix}
h_1 \\
h_2
\end{bmatrix}
\right)
\coloneq H_1
\end{equation}

\begin{equation}
F(H_1 W_{12}) = 
F\left(
\begin{bmatrix}
h_1 \\
h_2
\end{bmatrix}
\begin{bmatrix}
w_{31} & w_{32} \\
w_{41} & 0 \\
0 & w_{52}
\end{bmatrix}
\right)
=
F\left(
\begin{bmatrix}
h_3 \\
h_4 \\
h_5
\end{bmatrix}
\right)
\coloneq  H_2
\end{equation}

\begin{equation}
H_2' = H_1 \oplus H_2 =
\begin{bmatrix}
h_1 \\
h_2 \\
h_3 \\
h_4 \\
h_5
\end{bmatrix}
\end{equation}

\begin{equation}
F\left(H_2'W_{2O}\right)=
F\left(
\begin{bmatrix}
h_1 \\
h_2 \\
h_3 \\
h_4 \\
h_5
\end{bmatrix}
\begin{bmatrix}
0 & w_{2O} & w_{3O} & w_{4O} & w_{5O}
\end{bmatrix}
\right)
= O
\end{equation}

\subsection{Instantiating a matrix}

We will also present the instantiation of the $W_{12}$ matrix from this example according to the PropNEAT algorithm. The steps are to instantiate with zeros, and then to replace the weights with the connection weights from the genome. We will denote $C_{ij}$ to be a connection gene between nodes $i$ and $j$. The matching genome has the following connection genes: $[C_{1a}, C_{2a}, C_{1b}, C_{2b}, C_{2c}, C_{31}, C_{41}, C_{32}, C_{52}, C_{O2}, C_{O3}, C_{O4}, C_{O5}]$. 

We first instantiate the matrix with zeros

\begin{equation}
W^*_{12} = 
\begin{bmatrix}
0 & 0 \\
0 & 0 \\
0 & 0
\end{bmatrix}
\end{equation}

Then, for each of our connection genes within this layer, we replace the zero with the relevant weight. This results in the following matrix

\begin{equation}
W_{12} =
\begin{bmatrix}
w_{31} & w_{32} \\
w_{41} & 0 \\
0 & w_{52}
\end{bmatrix}
\end{equation}

\subsection{Demonstration of equivalence to NEAT}

We will demonstrate that the computation for the output is equivalent between PropNEAT and the NEAT implementation. As this includes inputs from a concatenated layer, a standard layer, and has a disconnect with one of the nodes, we use this as an example to demonstrate the equivalence. We acknowledge that it is not a full proof, but hope that it can help clarify the relationship between the two methods.

We will use $f(x)$ to denote the activation function applied to a single output, rather than the matrix. 

The NEAT genome has the following genes: $[C_{1a}, C_{2a}, C_{1b}, C_{2b}, C_{2c}, C_{31}, C_{41}, C_{32}, C_{52}, C_{O2}, C_{O3}, C_{O4}, C_{O5}]$. By the NEAT algorithm, the output node is calculated the following way.
\begin{align*}
O &= f(\Sigma_{k=1}^{j}w_{Ok}h_k) \quad \forall j: \exists C_{Oj} \\
  &= f(0 \cdot h_1 + w_{O2}h_2 + w_{O3}h_3 + w_{O4}h_4 + w_{O5}h_5 ) \\
  &= F\left(
\begin{bmatrix}
h_1 \\
h_2 \\
h_3 \\
h_4 \\
h_5
\end{bmatrix}
\begin{bmatrix}
0 & w_{2O} & w_{3O} & w_{4O} & w_{5O}
\end{bmatrix}
\right)
\end{align*}

This example covers the different cases of how nodes are connected, and generalises to layers with different numbers of nodes.

\end{document}